\documentclass[runningheads]{llncs}
\usepackage[T1]{fontenc}
\usepackage{hyperref}
\usepackage{graphicx,verbatim}
\usepackage{tabularx}
\usepackage{multirow}
\usepackage{booktabs}
\usepackage{amsmath}
\usepackage{amssymb}
\newcolumntype{A}{l}
\newcolumntype{B}{c}
\newcolumntype{Y}{>{\centering\arraybackslash}X}
\newcolumntype{S}{>{\centering\arraybackslash\hsize=0.5\hsize}X}
\newcolumntype{M}{>{\hsize=2.0\hsize}X}

\usepackage[dvipsnames]{xcolor}
\usepackage{enumerate}
\begin{document}
\title{LiteTracker: Leveraging Temporal Causality for Accurate Low-latency Tissue Tracking}
\titlerunning{LiteTracker} % Shorter title for the content page
\author{
Mert Asim Karaoglu\inst{1,2} \and
Wenbo Ji\inst{1,2} \and
Ahmed Abbas\inst{1} \and 
Nassir Navab \inst{2} \and
Benjamin Busam\inst{2} \and
Alexander Ladikos\inst{1}
}

\authorrunning{M.A. Karaoglu et al.}
\institute{
ImFusion \and
Technical University of Munich
}

% \author{Anonymized Authors}  %% Added for anonymized MICCAI 2025 submission
% \authorrunning{Anonymized Author et al.}
% \institute{Anonymized Affiliations \\
%     \email{email@anonymized.com}}

\maketitle              % typeset the header of the contribution
\begin{abstract}
Tissue tracking plays a critical role in various surgical navigation and extended reality (XR) applications.
While current methods trained on large synthetic datasets achieve high tracking accuracy and generalize well to endoscopic scenes, their runtime performances fail to meet the low-latency requirements necessary for real-time surgical applications.  
To address this limitation, we propose LiteTracker, a low-latency method for tissue tracking in endoscopic video streams.  
LiteTracker builds on a state-of-the-art long-term point tracking method, and introduces a set of training-free runtime optimizations.  
These optimizations enable online, frame-by-frame tracking by leveraging a temporal memory buffer for efficient feature re-use and utilizing prior motion for accurate track initialization.
LiteTracker demonstrates significant runtime improvements being around $7\times$ faster than its predecessor and $2\times$ than the state-of-the-art. 
Beyond its primary focus on efficiency, LiteTracker delivers high-accuracy tracking and occlusion prediction, performing competitively on both the STIR and SuPer datasets.  
We believe LiteTracker is an important step toward low-latency tissue tracking for real-time surgical applications in the operating room.
Our code is publicly available at \href{https://github.com/ImFusionGmbH/lite-tracker}{https://github.com/ImFusionGmbH/lite-tracker}.

% Authors must provide keywords and are not allowed to remove this Keyword section.

\end{abstract}
\section{Introduction}
Accurate tissue tracking is an essential component for a wide range of surgical robotics and extended reality (XR) applications~\cite{schmidt2024tracking,teufel2024oneslam,liang2024real,shinde2024jiggle}.
Endoscopic videos present a unique set of challenges for this task including strong non-rigid deformations, self-occlusions, viewpoint changes, abrupt camera motions, and further occlusions caused by surgical tools~\cite{karaoglu2024ride}. 
In addition to aforementioned challenges, employed algorithms must meet strict low-latency requirements due to the real-time demands of intraoperative applications. 

\begin{figure}[t]
    \centering
    \includegraphics[width=\textwidth]{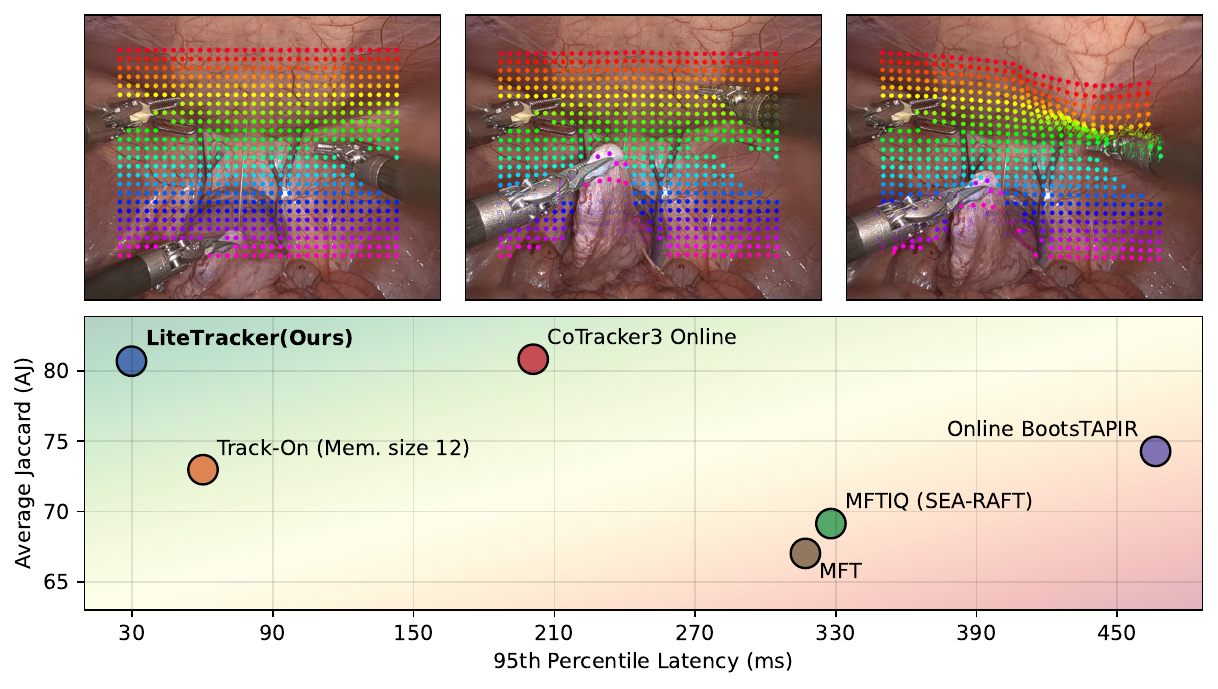}
    % \fbox{\rule{0pt}{4cm} \rule{11cm}{0pt}} % 6cm wide, 4cm high
    \caption{(Top) Demonstration of LiteTracker on a video from STIR Challenge 2024 dataset~\cite{schmidt2025point}.
    (Bottom) Latency and tracking accuracy comparison. Average Jaccard (AJ) metric is computed on SuPer dataset~\cite{li2020super}.
    Latencies showcase the 95th percentile of inference step measurements for 1,024 points initalized at the first frame of a video with 615 frames.
    LiteTracker is approximately $2\times$ faster than the fastest evaluated method, Track-On~\cite{aydemir2025track}, and $7\times$ than its predecessor, CoTracker3~\cite{karaev2024cotracker3}, while exhibiting close to state-of-the-art tracking accuracy.
    }
    \label{fig:teaser}
\end{figure}

Prior methods widely explore frame-to-frame motion for tissue tracking.
Sparse~\cite{schmidt2022fast,schmidt2023sendd} and dense~\cite{ihler2020self} optical and scene flow-based methods have been developed to track tissue motion in 2D and 3D.  
While these approaches effectively capture short-term motion, their restricted temporal context prevents occlusion handling and makes them susceptible to drift over long sequences.  
This issue becomes particularly pronounced in endoscopic videos, where occlusions, viewpoint changes, and rapid camera motions frequently disrupt tracking~\cite{chen2024mfst}.

Building on the foundations of Particle Videos~\cite{sand2008particle}, recent years have seen extensive research into long-term point tracking using both end-to-end~\cite{harley2022particle,doersch2024bootstap,karaev2024cotracker3,aydemir2025track} and hybrid~\cite{neoral2024mft,serych2024mftiq,chen2024mfst} approaches.
Compared to frame-to-frame methods, these models leverage a longer temporal context, improving tracking accuracy, enhancing robustness against drift, and handling occlusions more effectively.  
Trained on large realistic synthetic datasets and, in certain cases, fine-tuned on real world videos with weakly supervised training~\cite{doersch2024bootstap,karaev2024cotracker3,chen2024mfst}, they show successful generalization to endoscopic scenes~\cite{teufel2024oneslam,chen2024mfst}.
However, their low inference speed hinder their suitability for real-time applications.  

% Recently, A-MFST~\cite{chen2024mfst} extends the MFT~\cite{neoral2024mft} architecture and applies it for surgical tissue tracking, further demonstrating its effectiveness in this domain.  
% However, like MFT, A-MFST requires all tracking points to be initialized in the same frame, as adding new points later significantly degrades runtime performance.  

To this end, we introduce LiteTracker, a low-latency tissue tracking method tailored for real-time surgical applications.
Building upon CoTracker3~\cite{karaev2024cotracker3}, a state-of-the-art long-term point tracking approach, LiteTracker adopts a point-based representation and incorporates a set of training-free runtime optimizations.  

These enhancements enable LiteTracker to be approximately $7$ times faster than its predecessor~\cite{karaev2024cotracker3} and $2$ times than the fastest method evaluated in our experiments~\cite{aydemir2025track}, as shown in Fig.~\ref{fig:teaser}.

In addition to its efficiency gains, LiteTracker demonstrates strong tracking and occlusion prediction performance, ranking competitively on both the STIR Challenge 2024~\cite{schmidt2025point} and SuPer~\cite{li2020super} datasets.  

Our contributions are summarized as follows:
\begin{enumerate}
    \item A temporal memory buffer for caching previously computed features, enabling efficient per-frame predictions.
    \item Exponential moving average flow initialization for fast and robust tracking convergence without iterative updates.
    \item A low-latency, state-of-the-art tissue tracking method designed for integration as a building block into real-time surgical applications.
\end{enumerate}  

\section{Methodology}
We build LiteTracker using CoTracker3~\cite{karaev2024cotracker3} as our foundation due to its high tracking accuracy and strong generalizibility to endoscopic scenes.
In this section, we first give an overview of the CoTracker3 architecture and next, introduce our method, LiteTracker.

\subsection{Overview of CoTracker3}
CoTracker3's~\cite{karaev2024cotracker3} online model offers an end-to-end, transformer-based architecture that processes a video stream in a sliding-window fashion.
% Given a video stream $\left(I_t\right)_{t=1}^T$ consisting of $T$ frames, CoTracker3 predicts the visibility score $V_t \in [0, 1]$, confidence score $C_t \in [0, 1]$ and location $P_t = \left(x_t, y_t\right) \in \mathbb{R}^2$ of each query point $Q = \left(t^q, x_{t^q}, y_{t^q}\right)$, initialized at frame $t^q$, for the frames coming after their initialization. 
Given a video stream $\left(I_t\right)_{t=1}^T$ consisting of $T$ frames, and a set of query points $Q$ containing queries $q = \left(t^q, x_{t^q}, y_{t^q}\right)$ initialized at frame $t^q$, CoTracker3 predicts the visibility score $V_t \in [0, 1]$, confidence score $C_t \in [0, 1]$ and location $P_t = \left(x_t, y_t\right) \in \mathbb{R}^2$ for each query $q$ for all frames $t > t_q$ coming after their initialization. 

For new frames, the initial location, visibility and confidence scores ($P_t^{\text{init}}$, $V_t^{\text{init}}$, $C_t^{\text{init}}$) are set to the estimated values of the previous frame.

%TODOAA: Elaborate 'for the frames coming after their initialization' 
% For the sake of simplicity, our notations indicate a single query point with $Q$ whereas, the model processes a set of $N$ query points in parallel that can be initialized individually throughout the video stream.
The model operates in a sliding-window fashion, with a window size, $T_W$, of $16$ frames and stride, $S$, of $8$ frames. The tracking algorithm begins with a 2D convolutional neural network (CNN) to extract multi-scale feature maps $\Phi_t$ of each frame in the window, spatially downsampled by factors of $\{4, 8, 12, 16\}$ w.r.t.\ the input resolution. These feature maps are used to extract multi-scale patch features $\phi_{t}^s \in \mathbb{R}^{7 \times 7 \times d}$ around point locations. In the case of a query point, patch features $\phi_{t^q}^s$ are extracted around the query location $P_{t^q}$ from $\Phi_{t^q}$. These multi-scale query features are compared with the point features constructing a 4D correlation tensor~\cite{cho2024local}, which is then passed through a multi-layer perceptron (MLP) to generate correlation features.
The correlation features, current point estimates, confidence and visibility scores along with spatio-temporal positional encoding is passed to a transformer-based iterative refinement module.
This module predicts the updates for confidence $\Delta C_t$, visibility $\Delta V_t$, and location $\Delta P_t$ for all frames in the window. 
In the next refinement iteration, the updated location estimates are used to re-extract the patch features $\phi_t^s$, and the correlation features are recomputed.  
This process is repeated, as defined in the original implementation, $L:=6$ times, with the final estimates serving as the predictions.

\subsection{LiteTracker: An Efficient Solution}

\begin{figure}[t]
    \centering
    \includegraphics[width=\textwidth]{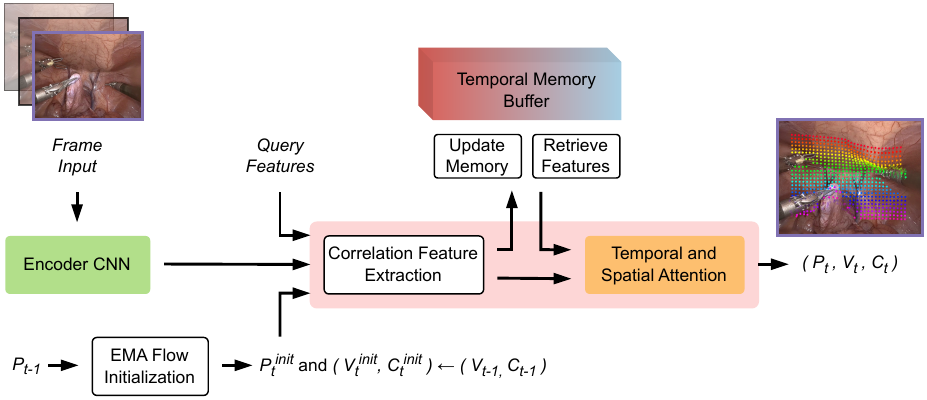}
    % \fbox{\rule{0pt}{4cm} \rule{11cm}{0pt}} % 6cm wide, 4cm high
    \caption{LiteTracker's architecture. Given a video stream and a set of query points, we extract feature maps for each new frame and compute correlation features between the queries and points initialized with exponential moving average flow (EMA flow).
    We store these correlation features in a temporal memory buffer for efficient re-use on subsequent frames. 
    Utilizing a transformer we propagate spatio-temporal information via attention mechanism to yield new point locations, visibility and confidence scores.}
    \label{fig:main}
\end{figure}

Despite its strong tracking performance, CoTracker3~\cite{karaev2024cotracker3} exhibits several efficiency limitations that hinder its deployment in real-time surgical applications.  
These bottlenecks arise primarily from its reliance on sliding-window processing and the iterative refinement module, both of which introduce computational overhead and latency.  
To this end, we introduce LiteTracker.
LiteTracker proposes a set of training-free runtime optimizations, see Fig.~\ref{fig:main}, to enable low-latency, accurate tissue tracking.

\subsubsection{Frame-by-frame Processing with Temporal Memory}
\label{sec:mem_buffer}
CoTracker3~\cite{karaev2024cotracker3} operates in a sliding-window manner requiring accumulation of a stride ($S:=8$) of frames before performing the next prediction.
A straight-forward way to perform frame-by-frame prediction is to set the stride $S$ to $1$ instead of $8$. 
This choice however, introduces a computational overhead since the processing of $T_W - S$ many frames need to be repeated multiple times.
To tackle this challenge, we propose maintaining a temporal memory buffer to cache prior computations. 
Concretely, we cache the correlation features since their computation is relatively expensive, involving pair-wise similarity measurement of multi-scale patch features. 
To this end we maintain a ring buffer, with a capacity $T_B$ of 16 frames, adhering to the original window size of CoTracker3.
The ring buffer operates in a first-in, first-out (FIFO) manner where the newly processed frames overwrite the oldest ones.
Our temporal memory buffer is expressly designed to allow the introduction of new query points at any time during tracking.
When additional points are queried in later frames, a set of proxy feature vectors are appended to the buffer.
The transformer architecture dynamically masks these proxies during temporal and spatial attention, ensuring that they do not influence existing tokens while preserving the causality of the pipeline.

\subsubsection{Track Initialization with Exponential Moving Average Flow}
\label{sec:ema_init}
Initializing the points on newly arrived frames by their prior locations, CoTracker3 originally performs multiple iterative refinement updates which includes computationally expensive components.
Since these iterations are executed sequentially, they impose a linear increase in runtime.  
Inspired by prior works in optical flow~\cite{wang2024sea}, we introduce a simple yet effective, training-free strategy to improve the convergence of the iterative refinement module by providing a more accurate location initialization.  
We achieve this using an exponential moving average flow (EMA flow) scheme, $F_t$, formulated as

\begin{equation}
    F_t = \alpha (P_{t-1} - P_{t-2}) + (1 - \alpha) F_{t-1},
\end{equation}

which is used to initialize the location for new frame as

\begin{equation}
    P_t^{\text{init}} = P_{t-1} + F_t.
\end{equation}

We empirically set the temporal smoothing factor $\alpha$, to $0.8$.
Exponential moving average flow initialization enables fast and robust trajectory extrapolation, allowing accurate location estimation in a single pass (i.e., $L:=1$) of the refinement module.

\begin{table}[t!]
	\caption{Comparison of long-term point tracking methods on STIR Challenge 2024~\cite{schmidt2025point} and SuPer~\cite{li2020super} datasets.
    $\delta^{\text{avg}}$ and AJ indicate respectively the average tracking accuracy and average Jaccard metrics across a set of pixel thresholds and OA indicates the occlusion accuracy metric. "*" designates that for that method, its score on the STIR dataset are taken from the official pre- and post-challenge submission results.
    Latencies showcase the 95th percentile of inference step (frame or window) measurements for 1,024 points initalized at the first frame of a video of 615 frames on an NVIDIA RTX 3090.
    }
	\label{tab:quantitative-comparison}
    \begin{tabularx}{\textwidth}{M Y S S S | Y}
\toprule
\multirow{2}{*}{Model} & \multirow{2}{*}{Input} & STIR & \multicolumn{2}{c}{SuPer} & \multirow{2}{*}{Latency (ms) $\downarrow$} \\
\cmidrule(lr){3-3} \cmidrule(lr){4-5}
& & $\delta^{\text{avg}}$ $\uparrow$ & AJ $\uparrow$ & OA $\uparrow$ & \\
\midrule
A-MFST*           & Frame  & 58.59 & --    & --    & -- \\
MFTIQ (ROMA)*     & Frame  & \underline{77.22} & 66.61 & 83.73 & 4355.11 \\
Online BootsTAPIR & Frame  & 68.59 & 74.27 & 93.63 & 466.38 \\
MFTIQ (SEA-RAFT)* & Frame  & 76.82 & 69.14 & 83.73 & 327.93 \\
MFT*              & Frame  & \textbf{77.62} & 67.02 & 83.73 & 317.05 \\
CoTracker3 Online & Win.   & 75.24 & \textbf{80.82} & \underline{96.96} & 200.98 \\
Track-On (48)     & Frame  & 74.44 & 70.22 & 83.92 & 74.80 \\
Track-On (12)     & Frame  & 72.74 & 72.97 & 86.86 & \underline{60.18} \\
\midrule
\textbf{LiteTracker (Ours)} & Frame & 75.81 & \underline{80.68} & \textbf{97.45} & \textbf{29.67} \\
\bottomrule
\end{tabularx}

\end{table}

\section{Experiments}
\subsection{Implementation Details and Experimental Setup}
\label{sec:implementation}
In LiteTracker, we do not perform any training and directly use the pretrained model weights from CoTracker3 online~\cite{karaev2024cotracker3}, which were obtained by training on TAP-Vid Kubric~\cite{doersch2022tap} and then fine-tuning on a curated set of unlabeled real videos through weak supervision. 

We conduct all experiments on a single NVIDIA RTX 3090 GPU using PyTorch~\cite{paszke2019pytorch}, with half-precision enabled for all methods from the literature where recommended by the authors.  
For a fair comparison, we empirically found that CoTracker3 achieves its highest performance on tissue tracking when the number of iterative refinement steps, $L$, is set to 4 instead of the original value of 6.  
Therefore, for CoTracker3, we set $L := 4$ in all evaluations. 
Following the previous benchmarks~\cite{doersch2022tap}, we utilize the suggested input size of each method for inference, in the case of LiteTracker, $512 \times 384$, and rescale the estimated tracks to the benchmark's evaluation size.

\subsection{Evaluation}
\subsubsection{Datasets}
We compare LiteTracker with the state-of-the-art long-term point tracking methods on two different tissue tracking benchmarks~\cite{schmidt2025point,li2020super}.

For the evaluation on STIR dataset~\cite{schmidt2024surgical}, we use its subset utilized for STIR Challenge 2024~\cite{schmidt2025point}.
This dataset consists of 60 sequences of different lengths, the longest one being up to 4 minutes. 
It includes sparse annotations, labeling the initial and end locations of tracked point in the initial and the last frame.
It utilizes the average tracking accuracy metric ($\delta^{\text{avg}}$) computed at the original image size, $1280 \times 1024$, with accuracy thresholds set to $\{4, 8, 16, 32, 64\}$ pixels.
Since it lacks the ground truth location and occlusion labels for the intermediate frames, this benchmark only measure the tracking performance.

We additionally evaluate LiteTracker on the SuPer~\cite{li2020super} dataset which consists of a single video of 522 frames, with manually annotated location and occlusion labels of 20 points for every $10^\text{th}$ frame.
Leveraging the ground truth occlusion labels, we employ the average Jaccard (AJ) and occlusion accuracy (OA) metrics to measure the tracking and visibility prediction performance.
Following their original implementation~\cite{doersch2022tap}, we compute AJ on the image size of $256 \times 256$ with accuracy thresholds set to $\{1, 2, 4, 8, 16\}$.

\subsubsection{Results}
We present our results in Tab.~\ref{tab:quantitative-comparison}.  
For the STIR dataset, we report the official pre- and post-challenge submissions, denoted with "*", as these methods did not have direct access to the dataset, which may have limited their ability to fine-tune their models for better performance.  
For the remaining methods, we use publicly available implementations where possible; otherwise, unavailable results are marked with "--".  

As highlighted in Tab.~\ref{tab:quantitative-comparison}, LiteTracker delivers competitive tracking performance on the STIR dataset, significantly outperforming the previous fastest state-of-the-art method that we evaluate on, Track-On~\cite{aydemir2025track}.  
On the SuPer dataset, LiteTracker ranks second in tracking accuracy, just behind CoTracker3~\cite{karaev2024cotracker3}, while achieving the highest occlusion prediction score.  

Despite its high accuracy, LiteTracker sets a new state-of-the-art in runtime efficiency, achieving an inference latency of 29.67 ms, measured as the 95th percentile over a 615-frame video tracking 1,024 points.  
This represents a substantial improvement over existing methods, running approximately $2$ times faster than Track-On~\cite{aydemir2025track} and $7$ times faster than its predecessor, CoTracker3~\cite{karaev2024cotracker3}.
Moreover, when deploying CoTracker3 in a real-time application, the implicit latency introduced by frame accumulation for sliding-window processing must also be considered.  
For instance, with an input video stream running at a frame rate $R$ of 30 Hz, this results in an additional delay of $(S-1) \times 1000 \div R$, (233.33 ms), leading to a total latency of 434.31 ms.  
This makes CoTracker3 approximately $16.6$ times slower than our method.  

\begin{figure}[t]
    \centering
    \includegraphics[width=\textwidth]{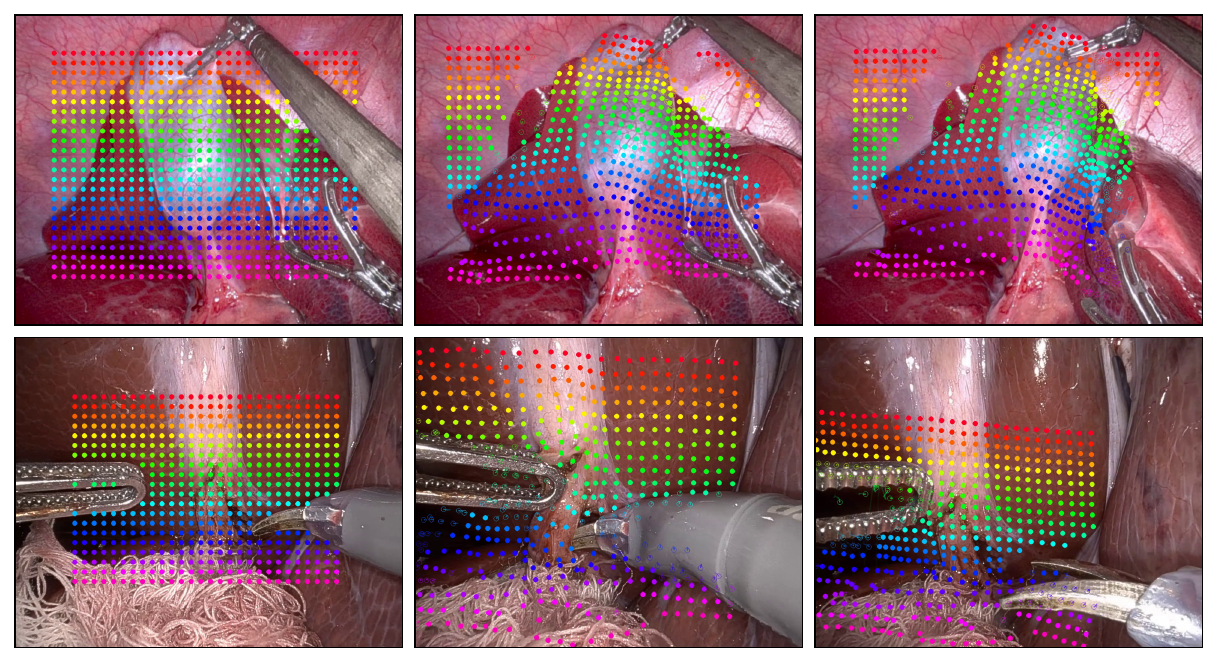}
    % \fbox{\rule{0pt}{4cm} \rule{11cm}{0pt}} % 6cm wide, 4cm high
    \caption{Qualitative results on video samples from the STIR Challenge 2024~\cite{schmidt2025point} (top) and StereoMIS~\cite{hayoz2023learning} (bottom) datasets. LiteTracker shows high tissue-tracking accuracy and occlusion handling under challenging deformations, tool interactions and perspective changes.}
    \label{fig:qualitative}
\end{figure}

Finally, Fig.~\ref{fig:qualitative} showcases LiteTracker’s accurate tissue tracking under viewpoint changes, occlusions, and deformations on video samples from STIR Challenge 2024~\cite{schmidt2025point} and StereoMIS~\cite{hayoz2023learning} datasets.

\subsubsection{Ablation Studies}

\begin{figure}[t]
    \centering
    \includegraphics[width=\textwidth]{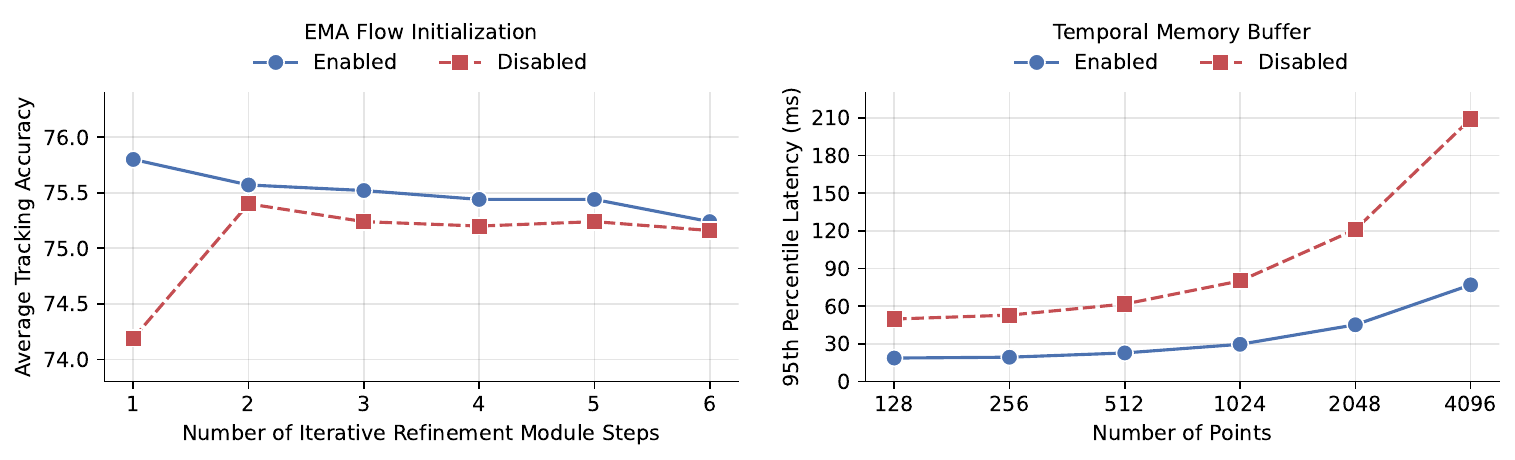}
    % \fbox{\rule{0pt}{4cm} \rule{11cm}{0pt}} % 6cm wide, 4cm high
    \caption{Ablation studies.
    (Left) Exponential moving average flow (EMA flow) initialization improves tracking convergence, leading to its highest average tracking accuracy on the STIR Challenge 2024 dataset~\cite{schmidt2025point} with a single pass through the iterative refinement module.
    (Right) Temporal memory buffer improves runtime efficiency by approximately $2.7$ times and provides low-latency inference during frame-by-frame tracking.
    }
    \label{fig:ablation}
\end{figure}

We conduct further experiments to analyze the impact of our contributions, providing their results in Fig.~\ref{fig:ablation}.

In our first study, we evaluate the EMA flow initialization's contribution to the overall tracking accuracy with respect to the varying number of refinement iterations.
Our results, collected on the STIR Challenge 2024 dataset~\cite{schmidt2025point}, showcase that our EMA flow initialization already yields the model's highest tracking accuracy with a single pass through the iterative refinement module.
We believe that our initialization scheme mimics temporal smoothness common to deformations in surgical scenes, there by providing a robust first estimate. 
Unlike CoTracker3, see Sec.\ref{sec:implementation}, further refinement iterations degrade the tracking accuracy of LiteTracker.
We believe this can be attributed to the use of pre-trained weights.

In our second study, we analyze the effect of the temporal memory buffer on the runtime performance, tracking different number of points.
Our results showcase that our caching mechanism consistently improves the runtime performance and results at approximately $2.7$ times faster inference speeds.
Furthermore, its compact size of the stored features results in negligible memory usage, $64$ MB for 1,024 points, allowing them to be kept on the GPU for fast access.

\section{Conclusion}

In this paper, we propose LiteTracker, a low-latency tissue tracking method designed for real-time surgical applications.  
Building on CoTracker3, a state-of-the-art long-term point tracking method, LiteTracker incorporates a set of training-free runtime optimizations enabling efficient and accurate frame-by-frame tissue tracking.
Our experimental results demonstrate that LiteTracker achieves substantial runtime improvements, running approximately $7\times$ faster than its predecessor and $2\times$ faster than the state-of-the-art, making it well-suited for real-time deployment.  
Furthermore, despite being designed for efficiency, LiteTracker maintains competitive tracking accuracy, coming close to the state-of-the-art on the SuPer and STIR datasets.
We believe LiteTracker presents a significant step toward practical usage of low-latency, accurate tissue tracking and opens new avenues for future research utilizing it as a building block for real-time surgical applications. 

\begin{comment}  %% removed for anonymized MICCAI 2025 submission.
    
    % The following acknowledgement and disclaimer sections should be removed for the double-blind review process.  
    % If and when your paper is accepted, reinsert the acknowledgement and the disclaimer clause in your final camera-ready version.

\begin{credits}
\subsubsection{\ackname} A bold run-in heading in small font size at the end of the paper is
used for general acknowledgments, for example: This study was funded
by X (grant number Y).

\subsubsection{\discintname}
It is now necessary to declare any competing interests or to specifically
state that the authors have no competing interests. Please place the
statement with a bold run-in heading in small font size beneath the
(optional) acknowledgments\footnote{If EquinOCS, our proceedings submission
system, is used, then the disclaimer can be provided directly in the system.},
for example: The authors have no competing interests to declare that are
relevant to the content of this article. Or: Author A has received research
grants from Company W. Author B has received a speaker honorarium from
Company X and owns stock in Company Y. Author C is a member of committee Z.
\end{credits}

\end{comment}
\bibliographystyle{splncs04}
\bibliography{LiteTracker}




% LNCS Homepage, \url{http://www.springer.com/lncs}, last accessed 2023/10/25
% \end{thebibliography}
\end{document}